\documentclass[acmtog, authorversion=True, nonacm=True]{acmart}

\usepackage{booktabs} 

\usepackage{multirow}
\usepackage{booktabs}

\begin{document}
\title{DeepJoin: Learning a Joint Occupancy, Signed Distance, and Normal Field Function for Shape Repair}

\author{Nikolas Lamb}
\orcid{0000-0002-6000-4658}
\affiliation{%
 \institution{Clarkson University}
 \city{Potsdam}
 \country{USA}
}
\email{lambne@clarkson.edu}

\author{Sean Banerjee}
\orcid{0000-0003-3085-056X}
\affiliation{%
 \institution{Clarkson University}
 \city{Potsdam}
 \country{USA}
}
\email{sbanerje@clarkson.edu}

\author{Natasha Kholgade Banerjee}
\orcid{0000-0001-7730-7754}
\affiliation{%
 \institution{Clarkson University}
 \city{Potsdam}
 \country{USA}
}
\email{nbanerje@clarkson.edu}

\renewcommand\shortauthors{Lamb, N. et al.}

\begin{abstract}
We introduce DeepJoin, an automated approach to generate high-resolution repairs for fractured shapes using deep neural networks. Existing approaches to perform automated shape repair operate exclusively on symmetric objects, require a complete proxy shape, or predict restoration shapes using low-resolution voxels which are too coarse for physical repair. We generate a high-resolution restoration shape by inferring a corresponding complete shape and a break surface from an input fractured shape. We present a novel implicit shape representation for fractured shape repair that combines the occupancy function, signed distance function, and normal field. We demonstrate repairs using our approach for synthetically fractured objects from ShapeNet, 3D scans from the Google Scanned Objects dataset, objects in the style of ancient Greek pottery from the QP Cultural Heritage dataset, and real fractured objects. We outperform three baseline approaches in terms of chamfer distance and normal consistency. Unlike existing approaches and restorations using subtraction, DeepJoin restorations do not exhibit surface artifacts and join closely to the fractured region of the fractured shape. Our code is available at: \url{https://github.com/Terascale-All-sensing-Research-Studio/DeepJoin}.
\end{abstract}

%
%


%
%


\begin{teaserfigure}
    \centering
      \includegraphics[width=\linewidth]{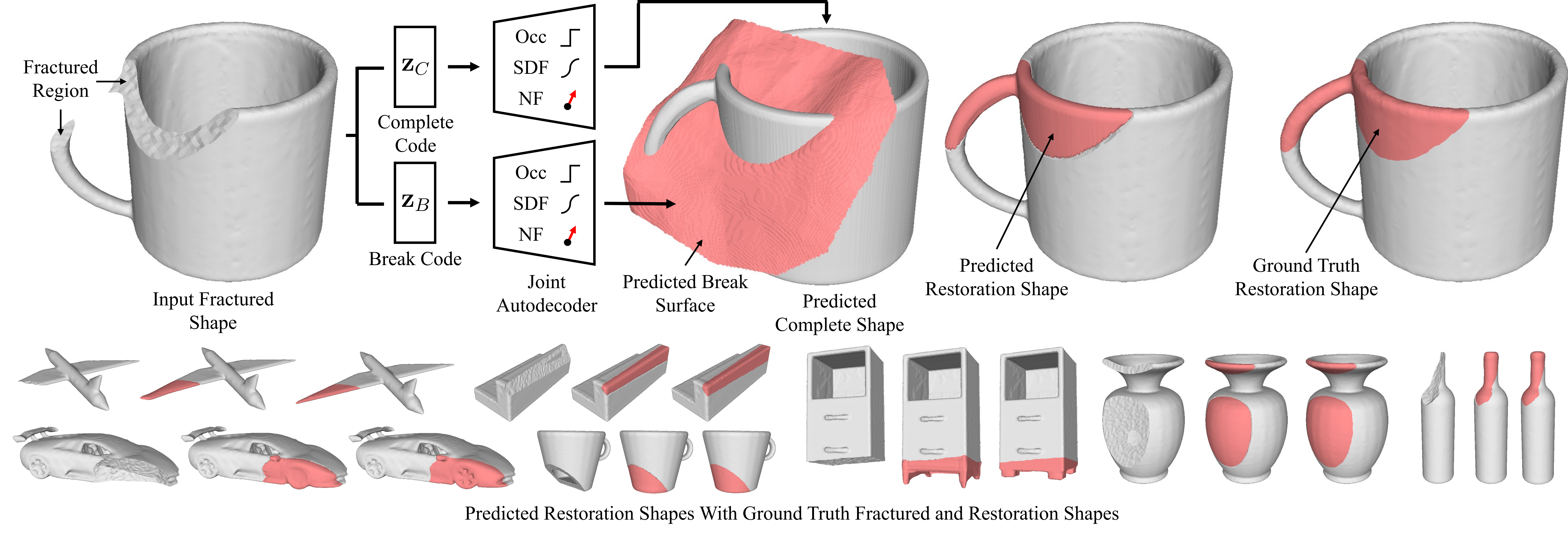}
      \caption{From an input fractured shape, DeepJoin estimates latent codes for a complete shape and break surface using occupancy (Occ), signed distance function (SDF), and normal field (NF) samples, and predicts a restoration shape to repair the input fractured shape using the estimated latent codes.}
    \label{fig__teaser}
\end{teaserfigure}

\maketitle

\section{Introduction}

\begin{figure}[t]
    \centering
      \includegraphics[width=\linewidth]{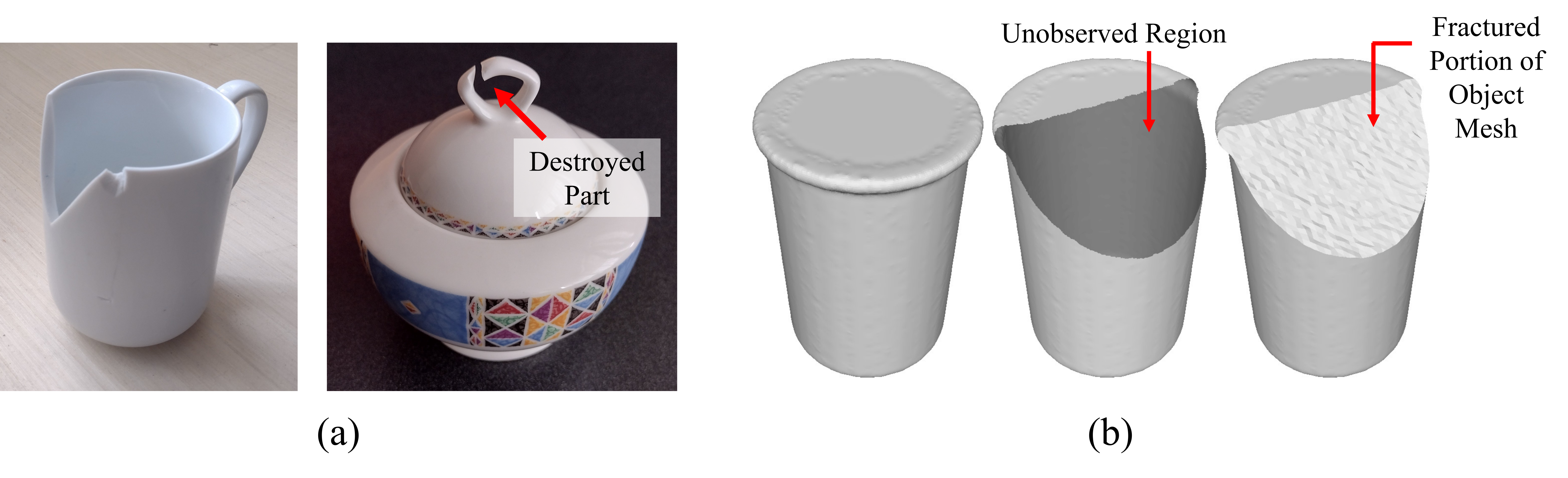}
      \caption{(a) A fractured mug and a fractured jar with a destroyed part. (b) A complete, partial, and fractured shape.}
    \label{fig__justify}
\end{figure}

Household objects often undergo damage, e.g. denting, part loss, weathering, or fracturing. Fractured objects may be reassembled if all of their parts can be found. However, in some cases parts are destroyed during the fracture, e.g. the sugar jar in Figure~\ref{fig__justify}(a), preventing reassembly of the object. Other damage, such as weathering, cannot be repaired using reassembly. Instead the object must be rebuilt using other materials. Users are likely to want to repair an object if the object has some sentimental or functional value. Repair is also necessary if the object is no longer manufactured. With the proliferation of 3D printing, the field of computational fabrication is ideally positioned to enable the repair of damaged objects by generating 3D printable repair parts.

Existing approaches that repair a single damaged object~\cite{scopigno20113d, seixas2018use, schilling2014reviving, antlej2011combining, rengier20103d} require modeling a repair part by hand using 3D design tools. While users may attempt to manually repair an object if the process is straightforward, repairing large or complex fractures, e.g. the cup in Figure~\ref{fig__justify}(a), is outside the scope of an average user. Most automated repair approaches, e.g. for aerospace components~\cite{gao2008integrated, zheng2006worn} and medical implants~\cite{harrysson2007custom, witek2016patient}, are domain-specific and unlikely to generalize. Other automated approaches require fractures to be symmetric to existing object parts~\cite{gregor2014towards, papaioannou2017reassembly}, or need to be fed a complete 3D proxy object~\cite{lamb2019automated}. Though 3D-ORGAN~\cite{hermoza20183d} operates directly on fractured objects without geometric constraints, it encodes objects in low-resolution voxel space, producing restorations that are too coarse for physical repair.

We present DeepJoin, an approach rooted in deep learning that infers a restoration shape from an input fractured shape by deconstructing the fractured shape into a corresponding complete shape and break surface, shown in Figure~\ref{fig__teaser}. Our approach is related to approaches that input a partial shape and perform shape completion by inferring a latent code to fit the partial shape observation~\cite{park2019deepsdf, sitzmann2020metasdf, tretschk2020patchnets, zheng2021deep, duggal2022mending, hao2020dualsdf, chen2019learning, yan2022shapeformer}. However, different from approaches that perform partial shape completion, our approach addresses the challenge that, unlike a partial shape that is a strict subset of the complete shape, as shown in Figure~\ref{fig__justify}(b), a fractured shape includes novel geometry at the fractured region that is not present in the complete shape.

To enable shape repair, our first contribution is a novel shape representation using a joint function that predicts three features\textemdash{}the occupancy function, the signed distance function (SDF), and the normal field. We find that when representing fractured shapes exclusively using the SDF, the most common implicit shape representation, deep networks struggle to accurately deconstruct the fractured shape into a complete shape and break surface. Our approach learns the occupancy function jointly with the SDF and partitions the fractured shape using the break occupancy, effectively treating estimation of the complete shape and break surface SDF as two partial completion tasks. The normal field (NF) represents surface normals for a shape. Our work uses the NF to capture the difference in surface properties between fractured and intact regions, in objects of materials such as ceramic or earthenware. Our second contribution is a set of loss functions for learning a representation of the complete shape and break surface using neural networks. Our third contribution is to use operations for constructive solid geometry (CSG) in occupancy and SDF space to formalize the dependence of fractured and restoration shapes on complete shapes and break surfaces in occupancy, SDF, and NF space. To generate a restoration mesh, we perform optimization over a fractured shape to obtain complete and break codes, predict the restoration SDF from the codes, and extract a mesh as the 0-level set. 

As no dataset of real fractured shapes currently exists, we synthetically fracture 24,208 meshes from the ShapeNet~\cite{shapenet2015} dataset and use them to validate our approach. We compare our approach to 3 baselines using existing shape completion approaches, i.e. DeepSDF~\cite{park2019deepsdf}, ONet~\cite{mescheder2019occupancy}, and ESSC~\cite{zhang2018efficient}. To demonstrate generalizability to cultural heritage objects and real 3D scans, we synthetically fracture and repair objects from the QP Cultural Heritage dataset~\cite{koutsoudis2009qp} and the Google Scanned Objects dataset~\cite{googlescannedobjects}. We also show restorations for physically fractured objects.

\section{Related Work}

\textit{Fractured Shape Restoration.}
Most existing approaches require a user to generate a repair part manually~\cite{scopigno20113d, seixas2018use, antlej2011combining, schilling2014reviving, rengier20103d}, which is out of the scope of the average user. Some approaches repair symmetric objects using reflection followed by subtraction~\cite{papaioannou2017reassembly, gregor2014towards}. These approaches cannot repair asymmetrical objects or objects that have undergone symmetric damage. The approach of Lamb et al.~\shortcite{lamb2019automated} requires a ground truth complete proxy shape as input, which may be unobtainable e.g. in the case of a rare or specialized object. 3D-ORGAN~\cite{hermoza20183d} performs shape completion from a fractured shape without requiring the shape to be symmetric. However, as 3D-ORGAN encodes shapes as $32^3$ resolution voxels, their restorations cannot accurately represent the fractured surface and cannot be closely joined to the fractured shape. DeepJoin infers a restoration shape directly without geometric constraints, and generates high-resolution restorations that fit closely to the fractured shape.

\textit{Partial Shape Completion.}
Though not directly related to our work, a large body of prior work has focused on performing shape completion from partial inputs using deep neural networks. Many approaches use point clouds~\cite{dai2017shape,han2017high,achlioptas2018learning, yuan2018pcn, sarmad2019rl, liu2020morphing, son2020saum, pan2021variational} due to their compactness. However, point clouds cannot intrinsically represent closed surfaces, which are necessary to generate repair parts that may be 3D printed. Approaches that predict meshes directly~\cite{yu2022part, groueix2018papier} struggle to reconstruct complex shapes~\cite{mescheder2019occupancy}, and cannot represent shapes of arbitrary topology. Voxel-based approaches~\cite{brock2016generative, sharma2016vconv, wu2016learning, smith2017improved, zhang2018efficient}, become computationally intractable at high resolutions. Though approaches have reduced the memory footprint of voxels using hierarchical models~\cite{dai2020sg, dai2018scancomplete} and sparse convolutions~\cite{dai2020sg, yi2021complete}, these approaches discretize the output space, rendering them incapable of representing high frequency geometry.

Many recent shape completion approaches encode shapes implicitly using the signed distance function (SDF)~\cite{xu2020ladybird, park2019deepsdf, sitzmann2020metasdf, yang2021deep, tretschk2020patchnets, zheng2021deep, lin2020sdf, ma2020neural, duggal2022mending, hao2020dualsdf, chabra2020deep}, the occupancy function~\cite{mescheder2019occupancy, chen2019learning, peng2020convolutional, jia2020learning, lionar2021dynamic, liao2018deep, yan2022shapeformer, yan2022implicit, sulzer2022deep, genova2020local, poursaeed2020coupling, chibane2020implicit}, or the unsigned distance function~\cite{tang2021sign, chibane2020neural, venkatesh2020dude}. DeepSDF~\cite{park2019deepsdf} introduced an autodecoder architecture that uses maximum \textit{a posteriori} estimation to perform shape completion by estimating a latent code to fit a set of SDF samples. Approaches have extended the autodecoder architecture by incorporating meta-learning~\cite{sitzmann2020metasdf}, encoding shapes at multiple resolutions~\cite{hao2020dualsdf}, and proposing incremental loss functions~\cite{duan2020curriculum}. Venkatesh et al.~\shortcite{venkatesh2020dude} propose a joint shape function using the unsigned distance and NF, and do not learn the occupancy function. We demonstrate that excluding the occupancy function results in inaccurate restoration shapes with a high chamfer distance. 

One could imagine an approach that identifies and removes the fractured region of the fractured shape to create a partial shape, applies a shape completion approach, and performs subtraction of the fractured shape from the complete shape to obtain a restoration shape. As we demonstrate in Section~\ref{sec:rescomp}, approaches based on subtraction produce artifacts on the surface of the fractured shape. Attempts to remove artifacts by discarding components that have a volume less than a threshold cannot eliminate connected artifacts. Our approach automatically generates restoration shapes that join closely to the fractured shape without producing artifacts.

\section{Representing Fractured Shapes}
\label{sec:rep}

\begin{figure}[t]
    \centering
      \includegraphics[width=0.97\linewidth]{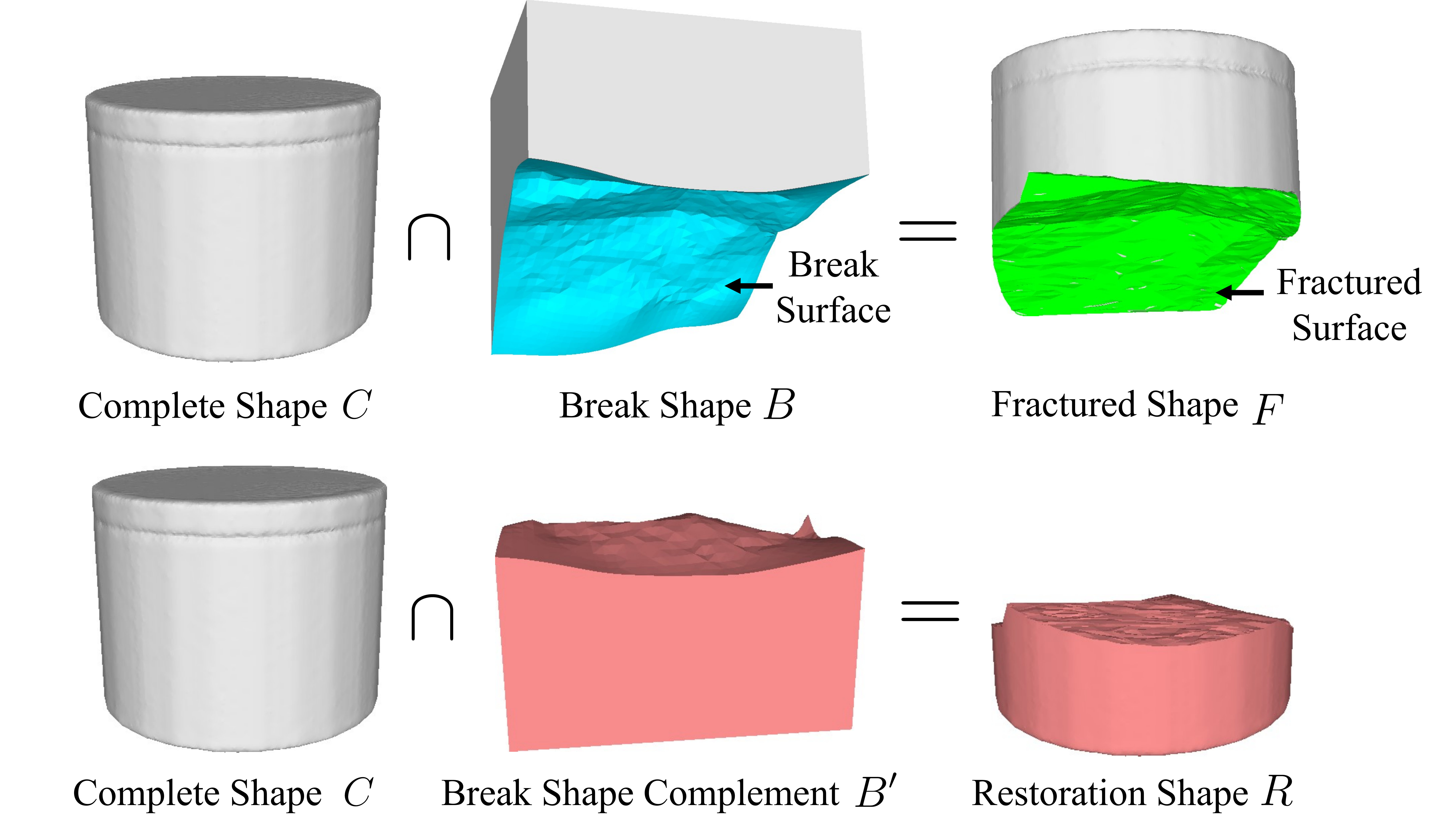}
      \caption{We express the fractured shape $F$ and the restoration shape $R$ as the intersection of the complete shape $C$ with the break shape $B$ and break shape complement $B'$ respectively. $B$ is separated from $B'$ by the break surface (blue). The fracture surface (green) is the surface shared by $B$ and $F$.}
    \label{fig__setrel}
\end{figure}

We represent complete, fractured, and restoration shapes as point sets $C$, $F$, and $R$ respectively. For a given shape $S \in \{ C, F, R \}$, we represent the occupancy of a point $\mathbf{x}$ as $o_S(\mathbf{x}) \in \{0,1\}$, where $o_S(\mathbf{x}) = 1$ if $\mathbf{x}$ is inside $S$ and 0 if $\mathbf{x}$ is outside or on the boundary of $S$. We exclude the boundary of $S$ to prevent a point from being inside multiple shapes simultaneously, e.g. inside $F$ and $R$. We define the break surface, shown in Figure~\ref{fig__setrel} in blue, as a 2D surface that intersects $F$ at the fractured region. As shown in Figure~\ref{fig__setrel}, we define the break shape, $B$, as the set of points on the same side of the break surface as the fractured shape. Theoretically the break shape $B$ has an infinite size. In practice, we limit the break shape to a occupy a unit cube. We define the fractured surface, shown in Figure~\ref{fig__setrel} in green, as the surface shared by the fractured shape and the break surface. We express the fractured shape as the intersection of the complete shape and the break shape, i.e. $F = C \cap B$. Similarly, we express the restoration shape as the intersection of the complete shape and the complement of the break shape, i.e. $R = C \cap B'$.

For a shape $S \in \{ C, F, R \}$, we define the SDF value of a point $\mathbf{x}$ as $s_S(\mathbf{x}) \in \mathbb{R}$, the signed distance from $\mathbf{x}$ to the surface of $S$. The value of $s_S(\mathbf{x})$ is negative inside the shape and positive outside. We define the NF value of a point $\mathbf{x}$ as $\mathbf{n}_S(\mathbf{x}) \in \mathbb{S}^3$, where $\mathbf{n}_S(\mathbf{x})$ is the normal vector of the closest point to $\mathbf{x}$ on the surface of $S$, and $\mathbb{S}^3$ is the unit sphere. We define the SDF $s_B(\mathbf{x})$ and the NF $\mathbf{n}_B(\mathbf{x})$ for the break shape as the signed distance to the fractured surface, shown in Figure~\ref{fig__setrel} in green, and the normal vector of the closest point to $\mathbf{x}$ on the fractured surface respectively. The value of $s_B(\mathbf{x})$ is negative on the fractured side and positive on the restoration side.

\begin{figure*}[t]
    \centering
      \includegraphics[width=\linewidth]{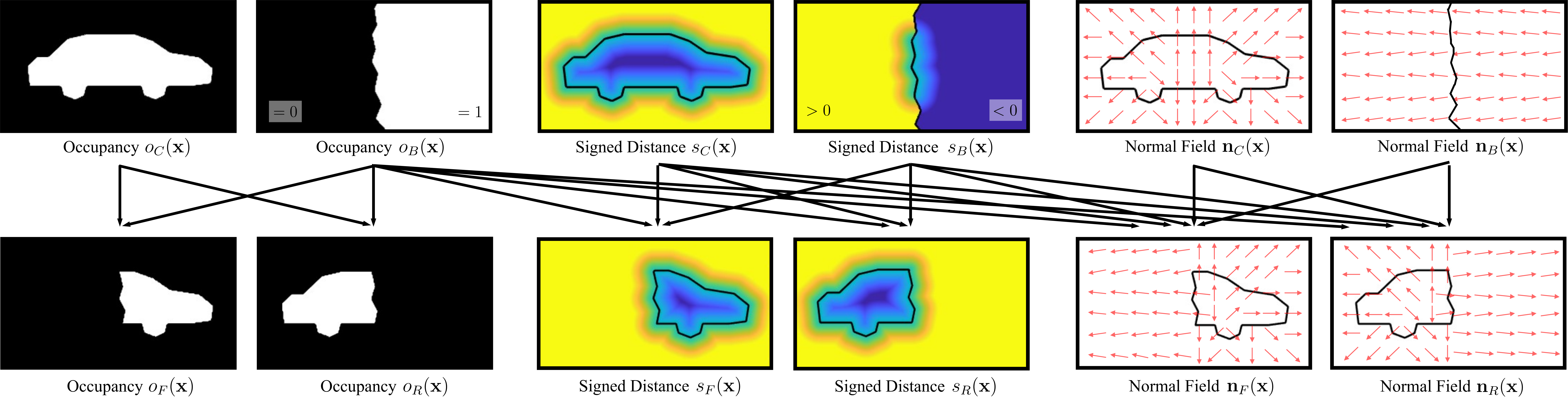}
      \caption{The fractured $o_F(\mathbf{x})$ and restoration occupancy $o_R(\mathbf{x})$ are obtained by carving out the complete shape with the break shape.
      For the SDF and NF, the fractured shape values $s_F(\mathbf{x})$ and $\mathbf{n}_F(\mathbf{x})$ are the same as the break shape for points outside and near the surface of the break shape, i.e., the same as $s_B(\mathbf{x})$ and $\mathbf{n}_B(\mathbf{x})$. The restoration shape values, $s_R(\mathbf{x})$ and $\mathbf{n}_R(\mathbf{x})$, are the opposite of the break shape for points inside and near the surface of the break shape.}
    \label{fig__concept}
\end{figure*}

The set definitions of $F$ and $R$ as $F = C \cap B$ and $R = C \cap B'$ imply that the point occupancies $o_F(\mathbf{x})$ and $o_R(\mathbf{x})$ for $F$ and $R$ can be expressed as a logical conjunction of occupancy in the complete shape, and occupancy in the break shape and the break shape complement, i.e. as $o_F(\mathbf{x}) = o_C(\mathbf{x}) \land o_B(\mathbf{x})$ and $o_R(\mathbf{x}) = o_C(\mathbf{x}) \land \neg o_B(\mathbf{x})$. The symbols $\land$ and $\neg$ represent the logical $\texttt{and}$ and $\texttt{not}$ operators. We relax these logical relationships to work with continuous values using the product T-norm~\cite{gupta1991theory}, as
\begin{align}
    & o_F(\mathbf{x}) = o_C(\mathbf{x})o_B(\mathbf{x}) \ \text{and} \label{eq:of} \\
    & o_R(\mathbf{x}) = o_C(\mathbf{x})(1- o_B(\mathbf{x})). \label{eq:or}
\end{align}
Figure~\ref{fig__concept} demonstrates the dependency of $o_F(\mathbf{x})$ and $o_R(\mathbf{x})$ on $o_C(\mathbf{x})$ and $o_B(\mathbf{x})$ given by Equations~\eqref{eq:of}~and~\eqref{eq:or}. In the figure, $o_F(\mathbf{x})$ and $o_R(\mathbf{x})$ can be seen as using the break shape and the inverse break shape to carve out the complete shape from a CSG perspective.

To compute the SDF value $s_F(\mathbf{x})$ for the fractured shape from the SDF values for the complete and break shapes, we express $s_F(\mathbf{x})$ as
\begin{align}
    s_F(\mathbf{x}) =
    \begin{cases}
      s_B(\mathbf{x}), & \text{if} \ o_B(\mathbf{x}) = 0 \ \text{or} \ s_B(\mathbf{x}) > s_C(\mathbf{x}) \\
      s_C(\mathbf{x}), & \text{otherwise}.
    \end{cases}
    \label{eq:sf}
\end{align}
Visually, Figure~\ref{fig__concept} demonstrates how the SDF values for $s_F(\mathbf{x})$ are obtained from the break shape for points outside the break shape and close to the break surface, i.e. when $s_B(\mathbf{x}) > s_C(\mathbf{x})$, following a CSG perspective in SDF space~\cite{breen20003d}, similar to the perspective for occupancy. We express the SDF value $s_R(\mathbf{x})$ for the restoration shape in terms of the complete and break shapes as
\begin{align}
    s_R(\mathbf{x}) =
    \begin{cases}
      -s_B(\mathbf{x}), & \text{if} \ o_B(\mathbf{x}) = 1 \ \text{or} \ -s_B(\mathbf{x}) > s_C(\mathbf{x}) \\
      s_C(\mathbf{x}), & \text{otherwise}.
    \end{cases}
    \label{eq:sr}
\end{align}
We negate values for the break shape in Equation~\eqref{eq:sr} compared to Equation~\eqref{eq:sf} as the restoration shape is on the opposite side of the break surface from the fractured shape.
Figure~\ref{fig__concept} illustrates how the SDF values for $s_R(\mathbf{x})$ are obtained from the inverted break shape for points outside the break shape and near the break surface where the SDF for the break shape is more positive than the complete shape, i.e. when $-s_B(\mathbf{x}) > s_C(\mathbf{x})$ holds.
Computing the SDF for the fractured and restoration shapes as a function of the occupancy for the break shape allows our approach to treat the estimation of the complete and break SDF as two partial shape completion problems, where the partial shapes are defined by the break occupancy.

Similar to the SDF, the NF value at a point $\mathbf{x}$ for a shape $S$ is given by the closest point on the surface of $S$ to $\mathbf{x}$. We use the same approach as in Equations~\eqref{eq:sf}~and~\eqref{eq:sr} to define the fractured NF $\mathbf{n}_F(\mathbf{x})$ and restoration NF $\mathbf{n}_R(\mathbf{x})$ in terms of $C$ and $B$, as
\begin{align}
    \mathbf{n}_F(\mathbf{x}) = & 
    \begin{cases}
      \mathbf{n}_B(\mathbf{x}), & \text{if} \ o_B(\mathbf{x}) = 0 \ \text{or} \ s_B(\mathbf{x}) > s_C(\mathbf{x}) \\
      \mathbf{n}_C(\mathbf{x}), & \text{otherwise}, \ \text{and} \label{eq:nf}
    \end{cases} \\
    \mathbf{n}_R(\mathbf{x}) = & 
    \begin{cases}
      -\mathbf{n}_B(\mathbf{x}), & \text{if} \ o_B(\mathbf{x}) = 1 \ \text{or} \ -s_B(\mathbf{x}) > s_C(\mathbf{x}) \\
      \mathbf{n}_C(\mathbf{x}), & \text{otherwise}. \label{eq:nr}
    \end{cases}
\end{align}
We negate the NF for the break shape in the definition of $\mathbf{n}_R(\mathbf{x})$ in Equation~\eqref{eq:nr}, as normals on the restoration shape at the fracture are oriented in the opposite direction of normals on the break surface.
Figure~\ref{fig__concept} shows that the normals for the fractured and break shape are the same outside of the break shape and near the break surface, i.e. where $s_B(\mathbf{x}) > s_C(\mathbf{x})$ holds. Normals for the restoration shape are also the opposite of the break shape inside the break and near the break surface, i.e. where $-s_B(\mathbf{x}) > s_C(\mathbf{x})$ holds.

\begin{figure}[t]
    \centering
      \includegraphics[width=0.9\linewidth]{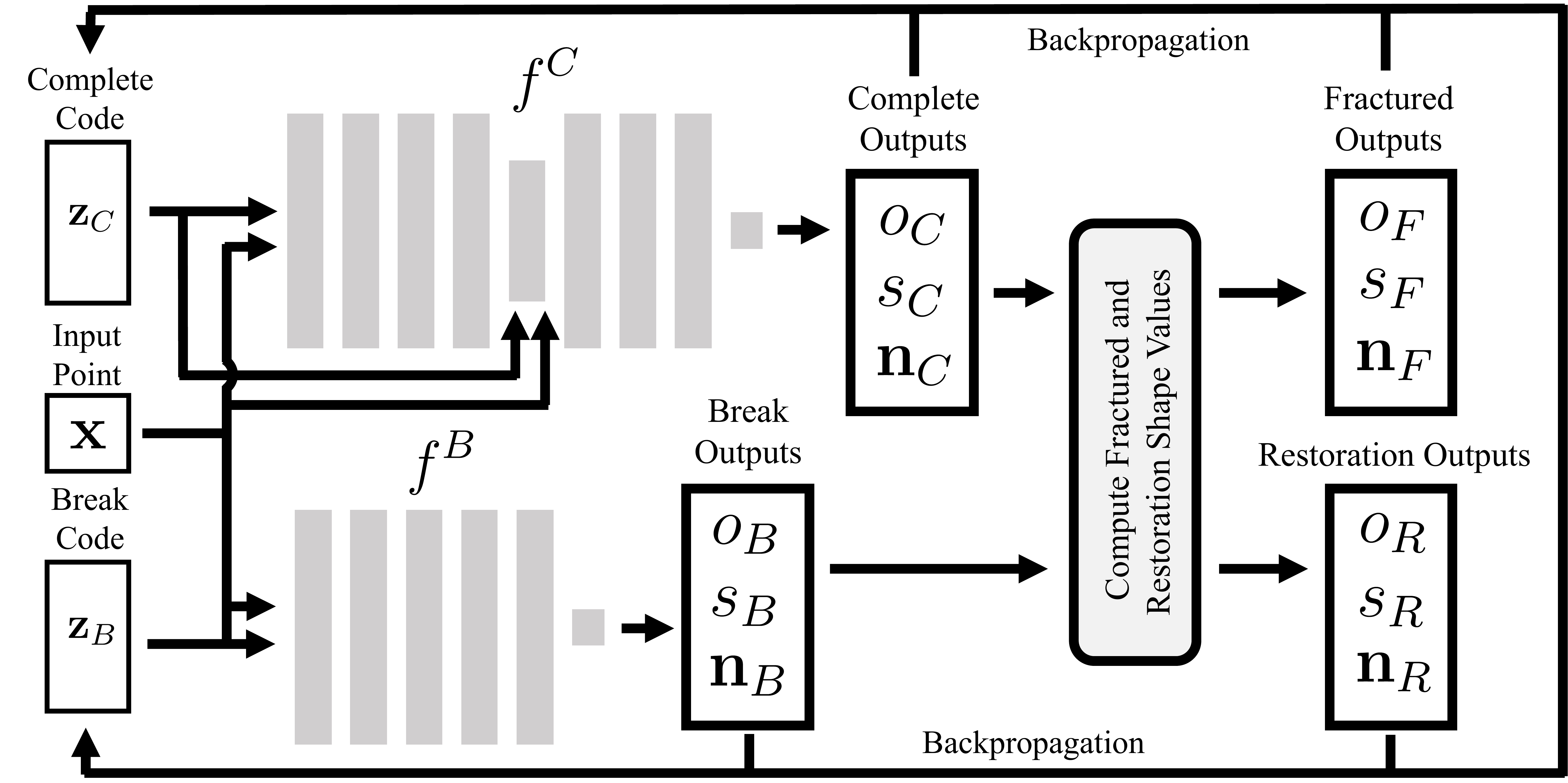}
      \caption{Networks $f^C$ and $f^B$ predict the occupancy, SDF, and NF values for $C$ and $B$ given an input point $\mathbf{x}$ and latent codes $\mathbf{z}_C$ and $\mathbf{z}_B$. We compute the occupancy, SDF, and NF values for $F$ and $R$ from the values for $C$ and $B$.}
    \label{fig__network}
\end{figure}

We represent the joint occupancy, SDF, and NF function $f^S$ for a shape $S \in \{C,B\}$ using a neural network based on the autodecoder architecture of DeepSDF~\cite{park2019deepsdf} as shown in Figure~\ref{fig__network}. We condition the autodecoder for $\mathcal{S}$ by associating $\mathcal{S}$ with a latent code $\mathbf{z}_{S} \in \mathbb{R}^p$ fed as input to the network, where $p$ is the size of the code. We use $p=128$ and $64$ for the complete and break shapes respectively. We use $f^S_{o}(\mathbf{z}_{S},\mathbf{x})$, $f^S_{s}(\mathbf{z}_{S},\mathbf{x})$, and $f^S_{\mathbf{n}}(\mathbf{z}_{S},\mathbf{x})$ to refer to the occupancy, SDF, and NF values predicted at point $\mathbf{x}$. 

\section{Network Optimization}

\label{subsec:train}
To train our networks, we use a dataset of sample shapes where each sample is a tuple $\{C, F, R, B\}$ corresponding to a complete, fractured, restoration, and break shape. We perform optimization over the network parameters, and over the complete and break shape codes. For a shape $S$, we use the notation $o_{S}(\mathbf{x})$, $s_{S}(\mathbf{x})$, and $\mathbf{n}_{S}(\mathbf{x})$ to represent the ground truth values for occupancy, the SDF, and NF respectively. During training, we optimize the loss
\begin{align}
    \mathcal{L}_\textrm{train} & = \textstyle\sum_{\mathbf{z}_C \in \mathcal{Z}_C, \mathbf{z}_B \in \mathcal{Z}_B} +  \mathcal{L}_{CB} + \mathcal{L}_F + 
    \mathcal{L}_R  + \lambda_\textrm{reg}\mathcal{L}_\textrm{reg},
\end{align}
where $\mathcal{Z}_C$ and $\mathcal{Z}_B$ are the sets of complete and break shape latent codes.
We define the loss $\mathcal{L}_{CB}$ for the complete and break shapes as
\begin{align}
    \mathcal{L}_{CB} & = \frac{1}{|\mathcal{X}|} \textstyle\sum_{S \in \{C, B\}} \textstyle\sum_{\mathbf{x} \in \mathcal{X}}
    \bigl(
        BCE(
            f^S_{o}(\mathbf{z}_{S},\mathbf{x}),
            o_{S}(\mathbf{x})
        ) + \nonumber \\
    &    \lambda_\textrm{s} ||
            f^S_{s}(\mathbf{z}_{S},\mathbf{x}) -
            s_{S}(\mathbf{x})
        ||_1
        + 
       \lambda_\textrm{n} ||
            f^S_{\mathbf{n}}(\mathbf{z}_{S},\mathbf{x}) -
            \mathbf{n}_{S}(\mathbf{x})
        ||_2
    \bigr),
\end{align}
where $BCE$ is the binary cross-entropy loss and $\mathcal{X}$ is a set of probing sample points. We define the fractured shape loss $\mathcal{L}_F$ as
\begin{align}
    \mathcal{L}_F = \frac{1}{|\mathcal{X}|} 
        \textstyle\sum_{\mathbf{x} \in \mathcal{X}}
        \bigl(
        & BCE\left(
                f^C_{o}(\mathbf{z}_{C},\mathbf{x}) f^B_{o}(\mathbf{z}_{B},\mathbf{x}),
                o_{F}(\mathbf{x}) 
            \right) + \nonumber \\
        & \lambda_\textrm{s} \mathcal{L}_{F_{s}} + 
        \lambda_\textrm{n} \mathcal{L}_{F_{\mathbf{n}}}
        \bigr).
\end{align}
The first argument to $BCE$, corresponds to the definition for fractured occupancy from Equation~\eqref{eq:of}. We represent the reconstruction error for the fractured shape SDF $\mathcal{L}_{F_{s}}$, as
\begin{align}
    \mathcal{L}_{F_{s}} =
    \begin{cases}
        ||
            f^B_{s}(\mathbf{z}_{B},\mathbf{x}) -
            s_{F}(\mathbf{x})
        ||_1 & 
        \text{if} \ f^B_{o}(\mathbf{z}_{B},\mathbf{x}) \leq \mu \ \text{or} \ \\
        & \ \ \ f^B_{s}(\mathbf{z}_{B},\mathbf{x}) > f^C_{s}(\mathbf{z}_{C},\mathbf{x}), \\
        ||
            f^C_{s}(\mathbf{z}_{C},\mathbf{x}) -
            s_{F}(\mathbf{x})
        ||_1 & \text{otherwise},
        \label{eq:losssf}
    \end{cases}
\end{align}
and the reconstruction error for the fractured shape NF $\mathcal{L}_{F_{n}}$, as
\begin{align}
    \mathcal{L}_{F_{\mathbf{n}}} =
    \begin{cases}
        ||
            f^B_{\mathbf{n}}(\mathbf{z}_{B},\mathbf{x}) -
            \mathbf{n}_{F}(\mathbf{x})
        ||_2 & 
        \text{if} \ f^B_{o}(\mathbf{z}_{B},\mathbf{x}) \leq \mu \ \text{or} \ \\
        & \ \ \ f^B_{s}(\mathbf{z}_{B},\mathbf{x}) > f^C_{s}(\mathbf{z}_{C},\mathbf{x}), \\
        ||
            f^C_{\mathbf{n}}(\mathbf{z}_{C},\mathbf{x}) -
            \mathbf{n}_{F}(\mathbf{x})
        ||_2 & \text{otherwise}.
        \label{eq:lossnf}
    \end{cases}
\end{align}
Equations~\eqref{eq:losssf}~and~\eqref{eq:lossnf} use the definitions for the fractured shape SDF and NF given in Equations~\eqref{eq:sf}~and~\eqref{eq:nf}. We use $\mu = 0.5$ as a threshold to determine if a point is inside the predicted break shape. $\mathcal{L}_R$, the loss for the restoration shape, is given as
\begin{align}
    \mathcal{L}_R = \frac{1}{|\mathcal{X}|} 
        \bigl(
        \textstyle\sum_{\mathbf{x} \in \mathcal{X}}
        & BCE\left(
                f^C_{o}(\mathbf{z}_{C},\mathbf{x}) (1-f^B_{o}(\mathbf{z}_{B},\mathbf{x})),
                o_{R}(\mathbf{x})
            \right) +  \nonumber \\
        & \lambda_\textrm{s} \mathcal{L}_{R_{s}} + 
        \lambda_\textrm{n} \mathcal{L}_{R_{\mathbf{n}}}
        \bigr).
        \label{eq:lossr}
\end{align}
The first argument to $BCE$ in Equation~\eqref{eq:lossr} corresponds to the definition for restoration shape occupancy given in Equation~\eqref{eq:or}. We give the restoration SDF reconstruction error $\mathcal{L}_{R_{s}}$ as 
\begin{align}
    \mathcal{L}_{R_{s}} =
    \begin{cases}
        ||
            -f^B_{s}(\mathbf{z}_{B},\mathbf{x}) -
            s_{R}(\mathbf{x})
        ||_1 & 
        \text{if} \ f^B_{o}(\mathbf{z}_{B},\mathbf{x}) > \mu \ \text{or} \ \\
        & \ \ \ -f^B_{s}(\mathbf{z}_{B},\mathbf{x}) > f^C_{s}(\mathbf{z}_{C},\mathbf{x}), \\
        ||
            f^C_{s}(\mathbf{z}_{C},\mathbf{x})
            -
            s_{R}(\mathbf{x})
        ||_1 & \text{otherwise},
    \end{cases} \label{eq:lossrs}
\end{align}
and the error for the restoration shape NF $\mathcal{L}_{R_{n}}$ as
\begin{align}
    \mathcal{L}_{R_{\mathbf{n}}} =
    \begin{cases}
        ||
            -f^B_{\mathbf{n}}(\mathbf{z}_{B},\mathbf{x}) -
            \mathbf{n}_{R}(\mathbf{x})
        ||_2 & 
        \text{if} \ f^B_{o}(\mathbf{z}_{B},\mathbf{x}) > \mu \ \text{or} \ \\
        & \ \ \ -f^B_{s}(\mathbf{z}_{B},\mathbf{x}) > f^C_{s}(\mathbf{z}_{C},\mathbf{x}), \\
        ||
            f^C_{\mathbf{n}}(\mathbf{z}_{C},\mathbf{x}) -
            \mathbf{n}_{R}(\mathbf{x})
        ||_2 & \text{otherwise},
    \end{cases}
    \label{eq:lossrn}
\end{align}
which use the definitions for the restoration shape SDF and NF from Equations~\eqref{eq:sr}~and~\eqref{eq:nr}.
We negate the SDF and NF value for the predicted break shape in Equations~\eqref{eq:lossrs}~and~\eqref{eq:lossrn} as the restoration lies on the opposite side of the break surface from the fractured shape. We define the regularization $\mathcal{L}_\textrm{reg}$ loss as
\begin{align}
    \mathcal{L}_\textrm{reg} = ||\mathbf{z}_{B}||_1 + ||\mathbf{z}_{C}||_1.
\end{align}
$\mathcal{L}_\textrm{reg}$ imposes a zero-mean Laplacian prior on the complete and break codes. We use $\lambda_s=1.0$, $\lambda_\mathbf{n}=1e-1$, and $\lambda_\textrm{reg}=1e-4$, i.e. the coefficients for the SDF, NF, and regularization losses respectively. We use the Adam optimizer~\cite{kingma2014adam}.

\section{Inferring Restoration Shapes}
At inference time our approach generates a restoration shape for a novel fractured shape by performing optimization over occupancy, SDF, and NF samples from the fractured shape to obtain complete and break shape codes. We use the codes to predict the SDF value for the restoration shape. During inference we optimize the loss
\begin{align}
    \mathcal{L}_\textrm{inf} & = \mathcal{L}_F +  \lambda_\textrm{reg}\mathcal{L}_\textrm{reg},
\end{align}
We find that the surface for the break shape estimated in occupancy space may deviate slightly from the surface in SDF space, causing artifacts if Equation~\eqref{eq:sr} is used to reconstruct the restoration mesh. To prevent artifacting, when generating the restoration mesh we obtain SDF values using the CSG equation for Boolean subtraction~\cite{breen20003d} in SDF space, i.e. $s_R(\mathbf{x}) = \max(f^C_{s}(\mathbf{z}_{C},\mathbf{x}), -f^B_{s}(\mathbf{z}_{B},\mathbf{x}))$, in place of Equation~\eqref{eq:sr}. To obtain a restoration mesh, we perform Marching Cubes~\cite{lorensen1987marching} on a $256^3$ grid of points.

\section{Data Processing and Datasets}

We evaluate our approach on four datasets.

\textit{ShapeNet.} We use 3 ShapeNet~\cite{shapenet2015} classes corresponding to commonly fractured objects, i.e. jars, bottles, and mugs, and 5 classes with more complex geometry, i.e. airplanes, chairs, cars, tables, and sofas. 
We train one network per ShapeNet class.

\textit{Google Scanned Objects dataset.} The dataset~\cite{googlescannedobjects} contains 3D scanned meshes of household objects such as shoes, pots, and plates. We train one network for the entire dataset.

\textit{QP Cultural Heritage dataset.} The dataset~\cite{koutsoudis2009qp} contains artist designed meshes in the style of Greek pottery. We use all meshes for testing on a network trained on ShapeNet jars.

\textit{Real Fractured Objects.} We fracture and scan 3 mugs and use 2 items from Lamb et al.~\shortcite{lamb2019automated}. We test them against networks trained on synthetically fractured ShapeNet mugs and jars respectively.

As the ShapeNet, Google, and QP datasets do not contain fractured meshes, we synthetically fracture meshes from these datasets using the fracturing approach described by Lamb et al.~\shortcite{lamb2021using}. To generate closed meshes we use the approach of Stutz and Geiger~\shortcite{Stutz2018ARXIV}. We normalize meshes so they occupy a unit cube. We perform a fracture retention test by fracturing each mesh using a randomized geometric primitive 15 times. If between 5\% and 20\% of the vertices of the mesh are not removed after 15 attempts we discard the mesh. As the bottles, jars, and mugs ShapeNet classes have less than 600 samples we fracture meshes from these classes 3, 3, and 10 times respectively. We fracture all other meshes once. We retain 24,208 out of 26,166 meshes from ShapeNet, 1,042 out of 1,298 meshes from the Google dataset, and 333 out of 408 meshes from the QP dataset.

Though meshes from the Google dataset are oriented upright they are not facing in a uniform direction. We augment the Google dataset by randomly rotating meshes around the ground plane normal by 90 degrees. We partition the ShapeNet and Google datasets using a 70\%/10\%/20\% train/validation/testing split. To obtain ground truth break surfaces we fit a thin-plate spline (TPS)~\cite{duchon1977splines} to the fractured region of each mesh such that the domain of the spline corresponds to a plane that is fitted to the fracture region vertices. We use the TPS to partition sample points into two sets, and denote the set that intersects with the fractured shape as the break shape. We discuss point our point sampling method in the supplementary.

\section{Results}
\label{sec:res}

We use the chamfer distance (CD), as defined by Park et al.~\shortcite{park2019deepsdf} and the normal consistency (NC), as defined by Mescheder et al.~\shortcite{mescheder2019occupancy} to evaluate the overall accuracy of predicted restoration shapes. In Section~\ref{sec:rescomp}, we compare our approach to methods based on subtraction from completed shapes. Restorations generated using subtraction tend to exhibit physically implausible surface artifacts. We contribute the non-fractured region error (NFRE) metric to evaluate the degree of surface artifacting. To compute the NFRE we sample $n$ points on the non-fractured region of the fractured shape and on the predicted and ground truth restoration shapes, and compute the percentage of points on the non-fractured region with a nearest neighbor on the predicted restoration that is closer than $\eta$ and a nearest neighbor on the ground truth restoration farther than $\eta$. We use $n=30,000$ and $\eta=0.02$. For success the NFRE and CD is low and the NC is high. In 1.2\% of cases, our approach generates an empty restoration. Where applicable, we show the non-empty percentage, (NE\%), i.e. the percentage of restorations generated.

\begin{figure}[t]
    \centering
      \includegraphics[width=\linewidth]{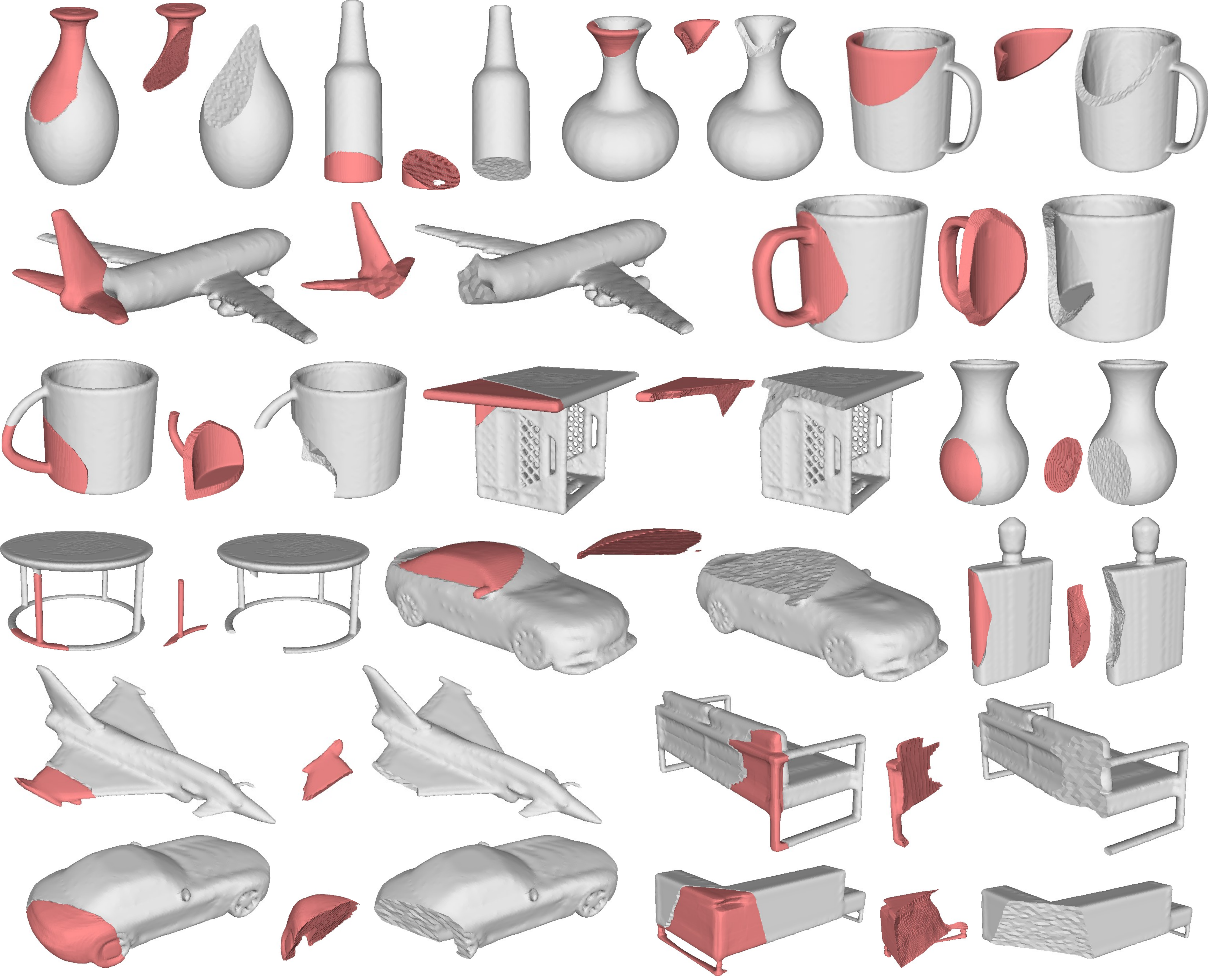}
      \caption{Predicted restoration shapes (red), joined to input fractured shapes (gray) and opened to show the fracture.}
    \label{fig__results}
\end{figure}

We show restoration shapes generated using our approach on the ShapeNet dataset in Figure~\ref{fig__results}. Our approach generates restoration shapes that join closely to the fractured region of the fractured shape and restore complex structures such as the handle of the mug on the right, and the thin table leg on the left of Figure~\ref{fig__results}. Restoration shapes keep with the style of the input shape, e.g. the connected feet of the sofa in the bottom right. Unlike approaches based on symmetry~\cite{gregor2014towards, papaioannou2017reassembly}, our approach repairs asymmetrical objects and objects with symmetrical fractures, such as the L-shaped sofa, airplane, beer bottle, vase, mugs, and both cars in Figure~\ref{fig__results}, and the mug, sofa, dresser, and bottle in Figure~\ref{fig__teaser}.

\begin{figure}[t]
    \centering
      \includegraphics[width=\linewidth]{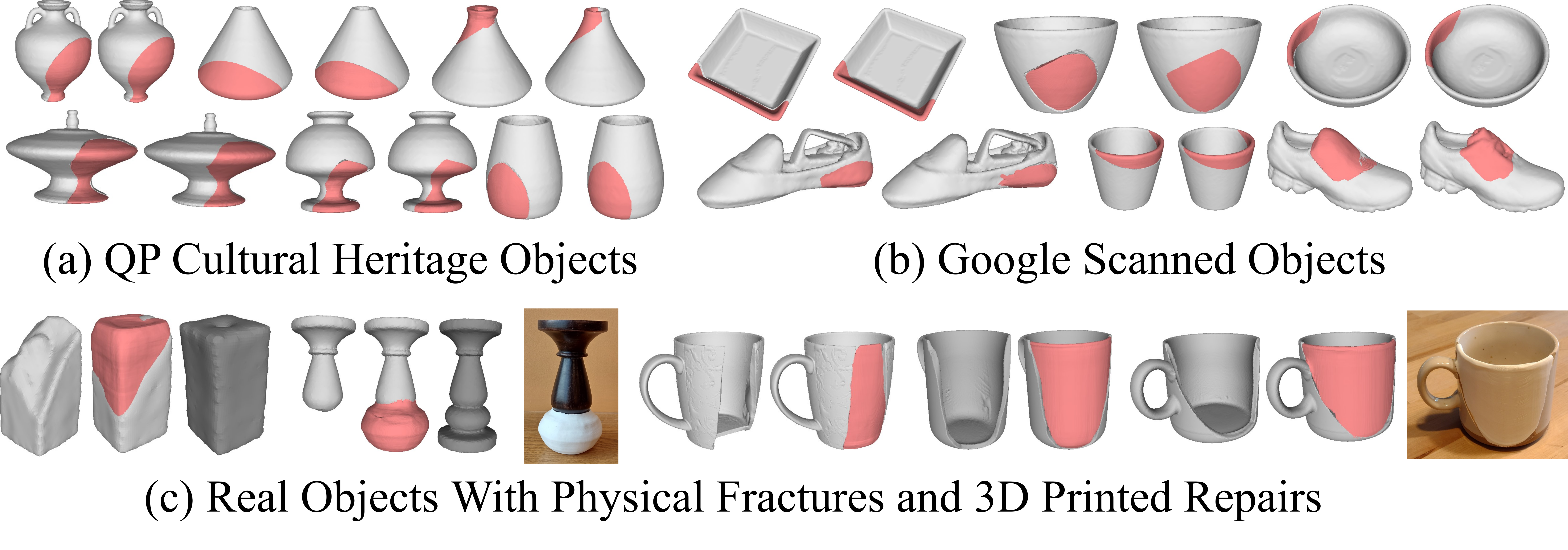}
      \caption{Predicted and ground truth restorations for synthetically fractured (a) objects in the style of ancient Greek pottery and (b) 3D scans of common objects. (c) Objects with real fractures and with two 3D printed restorations. Complete objects from Lamb et al.~\shortcite{lamb2019automated} are also shown in gray.}
    \label{fig__qp}
\end{figure}

In Figure~\ref{fig__qp} we show restorations generated for meshes from the QP Cultural Heritage dataset, 3D scans from the Google Scanned Objects dataset, and real fractured objects. Our approach obtains a CD of 0.117 on the Google Scanned Objects dataset and a CD of 0.144 on the QP dataset. Our approach is able to generate plausible restoration shapes for Greek pottery, as shown in Figure~\ref{fig__qp}(a), even having never been trained on objects from that time period. Though the Google Scanned Objects dataset is highly varied, our approach generates closely fitting repairs for simple household objects such as plates, bowls and cups, and for more complex objects, e.g. shoes, as shown in Figure~\ref{fig__qp}(b). Figure~\ref{fig__qp}(c) demonstrates that our approach generates feasible restoration shapes for real fractured objects, even when trained entirely on synthetic fractures. We 3D print a restoration part for the candlestick and the mug on the right. Though the base of the candlestick does not match the complete shape, the predicted restoration is physically plausible. For the mug, while small deviations in structure introduced by waterproofing and printer tolerances occur, the print provides a close fit, enabling repair.

\subsection{Ablation Study: Joint Function Modalities} \label{sec:resfunc}

We evaluate the impact of learning a joint function for multiple features by training our approach to represent fractured shapes using joint functions for the SDF alone, for occupancy alone (`Occ'), SDF and NF (`SDF+NF'), occupancy and SDF (`Occ+SDF'), and for all three features (`Occ+SDF+NF'). We show the CD and NE\% for each variation of our approach in Table~\ref{tab__mode}, over all restorations predicted by each approach. We do not show results for SDF as we find that the network is not able to learn a stable representation for the break shape and predicts no restorations. As shown in Table~\ref{tab__mode}, using occupancy gives a low CD of 0.099. However, it often fails to converge, with a NE\% of 89.4\%. Though SDF+NF shows a higher CD than Occ, it predicts restorations more often. Using Occ+SDF shows lower CD compared to Occ, and predicts more restorations than SDF+NF, with a NE\% of 98.9\%. Adding NF gives lowest CD and a similar NE\%, showing that NF improves restoration fidelity.

\setlength{\tabcolsep}{1.4pt}
\begin{table}[t!]
\footnotesize
\centering
\begin{tabular}{c|cccccccc|c|c}
\toprule
Method     & bottles        & cars           & chairs         & jars           & mugs           & planes         & sofas          & tables         & Mean           & NE\%          \\ \hline \hline
Occ        & 0.047          & 0.089          & 0.159          & 0.092          & 0.048          & 0.057          & 0.135          & 0.170          & 0.099          & 89.4          \\
SDF+NF     & 0.225          & 0.130          & 0.187          & 0.176          & 0.153          & 0.122          & 0.235          & 0.202          & 0.179          & 96.1          \\
Occ+SDF    & 0.042          & 0.023          & 0.127          & \textbf{0.071} & 0.028          & 0.043          & 0.101          & 0.159          & 0.074          & \textbf{98.9} \\
Occ+SDF+NF & \textbf{0.034} & \textbf{0.018} & \textbf{0.089} & 0.090          & \textbf{0.027} & \textbf{0.033} & \textbf{0.077} & \textbf{0.128} & \textbf{0.062} & 98.8    \\
\bottomrule
\end{tabular}
\caption{Chamfer (CD) and percentage of non-empty restorations (NE\%), using DeepJoin with different features. Best values are bolded.}
\label{tab__mode}
\end{table}

\subsection{Comparison: Fracture Removal and Shape Completion}
\label{sec:rescomp}

As no generalizable high-resolution shape repair approaches exist, we compare our approach to three baseline approaches that automatically remove the fractured region and perform shape completion. The baseline approaches use the existing shape completion approaches of DeepSDF~\cite{park2019deepsdf}, ONet~\cite{mescheder2019occupancy}, and ESSC~\cite{zhang2018efficient}. For each shape completion approach we create a partial input shape by removing the fractured region. To generate partial shapes, we train a point cloud classifier based on PointNet++~\cite{qi2017pointnet++} to classify the fractured region. Our classifier obtains a test accuracy of 97.1\%.

We train DeepSDF to reconstruct complete shapes from SDF point samples. We create a partial shape for inference by removing input sample points that have a nearest neighbor in the fractured region of the fractured shape identified by our classifier. We train ONet to reconstruct complete shapes from complete point clouds. We create a partial point cloud for inference by removing points that are classified as belonging to the fractured region identified by our classifier. 3D-ORGAN~\cite{hermoza20183d} generates low-resolution voxelized restoration shapes from fractured shapes. We find that the approach is unstable and does not converge during training. To provide a fair comparison to a voxel-based approach, we use ESSC~\cite{zhang2018efficient} to perform partial shape completion. We train ESSC to reconstruct complete voxel grids from partial voxel grids at $32^3$ spatial resolution to match the resolution of 3D-ORGAN. We use an input resolution of $128^3$ encoded with the flipped-truncated SDF (FTSDF)~\cite{song2017semantic}. During inference, we create a partial input by computing the FTSDF with respect to a mesh with the fractured region, identified by our classifier, removed.

For each baseline approach we obtain a restoration shape by subtracting the predicted complete shape from the input fractured shape in occupancy space. For DeepSDF and ONet we obtain a mesh using Marching Cubes for $256^3$ sample points. As restoration shapes must be closed meshes and not voxel grids, we generate a mesh for ESSC using Marching Cubes at $32^3$ resolution. To mitigate surface artifacts, for each of the baseline approaches we automatically remove connected components from the predicted restoration shape that have a volume less than $\delta$, where we use $\delta = 0.01$. 

\setlength{\tabcolsep}{1.8pt}
\begin{table}[t!]
\footnotesize
\centering
\begin{tabular}{c|c|cccccccc|c}
\toprule
Method                & Metric & bottles        & cars           & chairs         & jars           & mugs           & planes         & sofas          & tables         & Mean           \\ \hline \hline
\multirow{3}{*}{DSDF} & CD     & \textbf{0.024} & 0.025          & 0.109          & 0.134          & 0.043          & 0.034          & \textbf{0.076} & 0.129          & 0.072          \\
                      & NC     & \textbf{0.723} & 0.636          & 0.471          & 0.506          & 0.679          & 0.562          & \textbf{0.591} & 0.495          & 0.583          \\
                      & NFRE   & 0.084          & 0.158          & 0.316          & 0.217          & 0.092          & 0.125          & 0.215          & 0.238          & 0.181          \\ \hline
\multirow{3}{*}{ONet} & CD     & 0.103          & 0.122          & 0.160          & 0.133          & 0.150          & 0.121          & 0.148          & 0.190          & 0.141          \\
                      & NC     & 0.441          & 0.428          & 0.416          & 0.391          & 0.445          & 0.384          & 0.418          & 0.460          & 0.423          \\
                      & NFRE   & 0.829          & 0.753          & 0.518          & 0.538          & 0.523          & 0.787          & 0.649          & 0.473          & 0.634          \\ \hline
\multirow{3}{*}{ESSC} & CD     & 0.175          & 0.069          & 0.167          & \textbf{0.056} & 0.052          & 0.126          & 0.152          & \textbf{0.122} & 0.115          \\
                      & NC     & 0.108          & 0.388          & 0.153          & 0.414          & 0.481          & 0.138          & 0.251          & 0.377          & 0.289          \\
                      & NFRE   & 0.346          & 0.114          & 0.168          & \textbf{0.051} & 0.029          & 0.178          & 0.298          & \textbf{0.030} & 0.152          \\ \hline
\multirow{3}{*}{Ours} & CD     & 0.034          & \textbf{0.018} & \textbf{0.089} & 0.090          & \textbf{0.027} & \textbf{0.033} & 0.077          & 0.128          & \textbf{0.062} \\
                      & NC     & 0.687          & \textbf{0.749} & \textbf{0.567} & \textbf{0.558} & \textbf{0.783} & \textbf{0.660} & 0.537          & \textbf{0.505} & \textbf{0.631} \\
                      & NFRE   & \textbf{0.052} & \textbf{0.021} & \textbf{0.042} & 0.054          & \textbf{0.008} & \textbf{0.024} & \textbf{0.064} & 0.040          & \textbf{0.038} \\
\bottomrule
\end{tabular}
\caption{Chamfer distance (CD), normal consistency (NC) and non-fracture region error (NFRE) for baseline approaches and our approach. Best metric values are bolded. Mean is computed over class means.}
\label{tab__comp}
\end{table}

\begin{figure}[t]
    \centering
      \includegraphics[width=\linewidth]{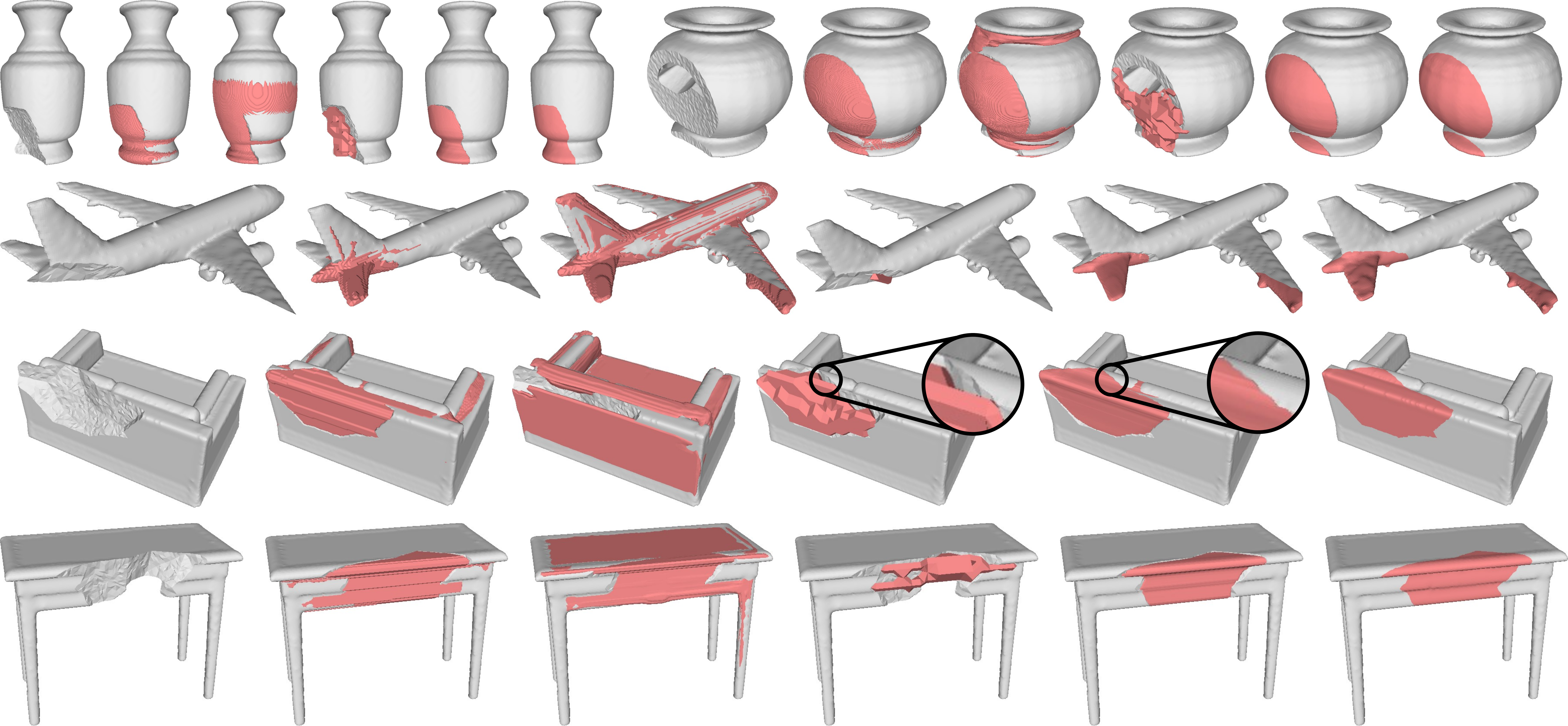}
      \caption{From left to right: input fractured shapes (gray) and restorations (red) from DeepSDF, ONet, ESSC, DeepJoin, and ground truth.}
    \label{fig__comp}
\end{figure}

Table~\ref{tab__comp} summarizes metrics over non-empty restorations predicted by our approach. Comparative approaches generate all non-empty restorations. DeepJoin outperforms baseline approaches in terms of the mean CD, NC, and NFRE. DeepSDF predicts restorations with a relatively low CD of 0.072 and high NC of 0.583 compared to the other baselines. However, as shown by the NFRE of 0.181 in Table~\ref{tab__comp}, restorations predicted by DeepSDF exhibit artifacts on the surface of the fractured shape e.g. for the bottle, sofa, and pot in Figure~\ref{fig__results}. Complete shapes predicted by ONet are less accurate than DeepSDF, resulting in restoration shapes with large artifacts that often cover most of the surface of the fractured shape, as shown in Figure~\ref{fig__comp}, and by the NFRE of 0.634. ESSC shows a the lowest mean CD for 2 classes, and often predicts restoration shapes in the correct location and without surface artifacts. However, the restoration shapes do not precisely match the ground truth, as shown by the NC of 0.289, do not join to the fractured shape, e.g. the cutaway in Figure~\ref{fig__comp}, and are too small to fully restore the fractured shape due to their coarse structure. DeepJoin generates accurate restoration shapes that do not exhibit artifacts, as shown in Figure~\ref{fig__comp}.

\section{Conclusion and Future Work}

In this paper we present DeepJoin, an approach to automatically generate restoration shapes from input fractured shapes by deconstructing the fractured shape into corresponding complete and break shapes. We contribute a novel method to encode a shape using a joint function for occupancy, signed distance, and NF and use this encoding to regress a high resolution restoration shape. Our approach overcomes the disadvantages of prior work using voxels for automated shape repair, and plays an important role in enabling consumer-driven object repair using additive manufacturing. 

\begin{figure}[t]
    \centering
      \includegraphics[width=\linewidth]{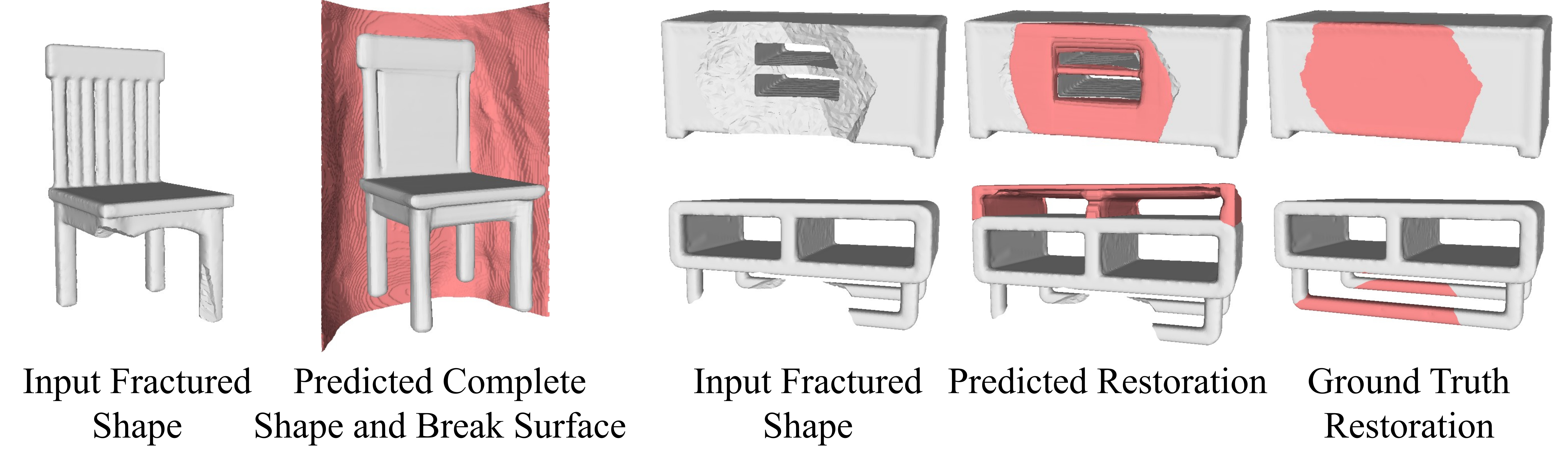}
      \caption{Left: The complete and break shape may not intersect. Right: Restorations may be small predicted in the wrong location.}
    \label{fig__limit}
\end{figure}

\textit{Limitations.} 
In 1.2\% of cases our approach fails to generate the restoration shape, e.g. on the left of Figure~\ref{fig__limit}, when the break surface is predicted away from the fractured shape. However, as our approach also predicts a complete shape, it enables fallback to generating a restoration using subtraction. As we represent break surfaces using thin-plate splines, our approach is unable to accurately represent concave breaks. In future work, we will explore using more generalizable representations for the break surface, such as non-uniform rational basis splines (NURBS)~\cite{piegl1996nurbs}. Our approach may predict a restoration shape that is smaller than the ground truth, e.g. the cabinet in Figure~\ref{fig__limit}, or may predict the restoration shape in the wrong location if the fractured region is small, e.g. the TV stand in Figure~\ref{fig__limit}. However, these repair parts can still be used to partially repair the fracture, e.g. for the table.

\textit{Future Work.}
Though we focus on single component fractures, future work can explore simultaneous estimation of inter-part transformations to facilitate multi-fracture assembly and repair. In future work, we will investigate automated repair deformation near the fracture surface to enable fine-precision joins of 3D prints for physical water-tightness. Though our fractured shapes simulate the surface roughness found in ceramic or earthenware objects, proliferation of our approach requires large-scale analysis of fracture across diverse materials. Our future work will contribute a comprehensive fracture dataset of 3D scans of physically fractured objects with accurate waterproofing, to facilitate data-driven study of fracture.

\bibliographystyle{ACM-Reference-Format}
\bibliography{acmart-bib}

\end{document}